\documentclass[11pt]{article}
\usepackage{colacl, qtree, epsfig}

\title{Compacting the Penn Treebank Grammar}

\author {Alexander Krotov \and Mark Hepple \and 
        Robert Gaizauskas \and Yorick Wilks
        \\Department of Computer Science, Sheffield
        University\\
        211 Portobello Street, Sheffield S1 4DP, UK\\
        {\tt \{alexk, hepple, robertg, yorick\}@dcs.shef.ac.uk}}

\begin{document}

\maketitle

\begin{abstract}

Treebanks, such as the Penn Treebank (PTB), offer a simple approach
to obtaining a broad coverage grammar: one can simply read the grammar
off the
parse trees in the treebank. While such a grammar 
is easy to obtain, a square-root rate of growth of the rule set with corpus 
size suggests that the derived grammar is far from
complete and that much more treebanked text would be required to obtain
a complete grammar, if one exists at some limit.  However, we offer an
alternative 
explanation in terms of the 
underspecification of structures within the treebank.
This hypothesis is
explored by applying an algorithm to {\em compact} the derived grammar
by eliminating redundant rules -- rules whose right hand sides can be
parsed by other rules. The size of the resulting compacted grammar,
which is significantly less than that of the full treebank grammar, is shown
to approach a limit.  However, such a compacted grammar does
not yield very good performance figures.  A version of the compaction
algorithm taking rule probabilities into account is proposed, which is
argued to be more linguistically motivated.  Combined
with simple thresholding, this method can be used
to give a 58\% reduction in grammar size without significant change 
in parsing performance, and can produce a 69\% reduction with 
some gain in recall, but a loss in precision.  

\end{abstract}

\newcommand{\myfigure}[1]{\input{figures/#1}}
\newcommand{\myepsffile}[1]{\epsffile{figures/#1.eps}}

\newcommand{\macfigs}{%
   \renewcommand{\myfigure}[1]{%
       \begin{center}
         \framebox[110mm]{\rule{0mm}{65mm}}
       \end{center}}
   \renewcommand{\myepsffile}[1]{\input{figures/}}}

\section{Introduction}

The Penn Treebank (PTB)
\cite{PTB94} has been used for a rather 
simple approach to deriving large grammars automatically: one where 
the grammar rules 
are simply `read off' the parse trees in the corpus, with each 
local subtree providing the left and right hand sides of 
a rule.  Charniak \cite{Charniak96} reports precision and recall
figures of around 80\% for a parser employing such a grammar.  In this
paper we show that the huge size of such a treebank grammar (see below) 
can be reduced in size without
appreciable loss in performance, and, in fact, an improvement in 
recall can be achieved.



Our approach can be generalised in terms of Data-Oriented Parsing
(DOP) methods (see \cite{Bonnema97}) with the
tree depth of 1.  However, the number of trees produced with a general
DOP method is so large that Bonnema \cite{Bonnema97} has to resort 
to restricting 
the tree depth, using a very domain-specific corpus such as ATIS or
OVIS, and parsing very short sentences of average length 4.74 words.
Our compaction algorithm can be 
extended for the use within the
DOP framework but, because of the huge size of the derived grammar
(see below), we chose to
use the simplest PCFG framework for our experiments.  

We are concerned with the nature of the rule
set extracted, and how it can be improved, with regard both to 
linguistic criteria and processing efficiency.
In what follows, we report the worrying observation that the 
growth of the rule set continues at a square root rate 
throughout processing of the entire treebank (suggesting, 
perhaps that the rule set is far from complete).  Our results 
are similar to those reported in \cite{KGW94}.
\footnote{For the complete investigation of the
grammar extracted from the Penn Treebank II see 
\cite{Gaizauskas95}}
We discuss an alternative possible source
of this rule growth phenomenon, {\it partial bracketting},
and suggest that it can be alleviated by {\it compaction}, where
rules that are redundant (in a sense to be defined) are eliminated
from the grammar.  

%

Our experiments on compacting a PTB treebank
grammar resulted in two major findings: one, that the grammar can 
at the limit be
compacted to about 7\% of its original size, and the rule number
growth of the compacted grammar stops at some point.  The other 
is that a 58\% reduction
can be achieved with no loss in parsing performance, whereas a 69\%
reduction yields a gain in recall, but a small loss in precision.  


This, we believe, gives
further support to the utility of treebank grammars and
to the compaction method.  For example, compaction 
methods can be applied within the DOP framework to reduce the number 
of trees.  Also, by partially lexicalising the rule extraction process
(i.e., by using some more frequent words as well as the part-of-speech
tags), we may be able to achieve parsing performance similar to the
best results in the field
obtained in \cite{Collins96}.



\section{Growth of the Rule Set}


One could investigate whether there is a finite grammar that should
account for any text within a class of related texts (i.e. a domain 
oriented
sub-grammar of English).  If there is, the number of extracted rules 
will approach a limit as more sentences are
processed, i.e. as the rule number approaches the size of such an 
underlying and finite grammar.




\begin{figure}

\epsfxsize=3in
\epsfysize=1.5in
\epsfbox{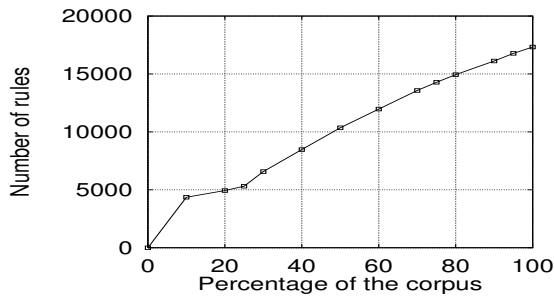}

\caption{Rule Set Growth for Penn Treebank II}
\label{fig.rule.growth.compare}
\end{figure}

We had hoped that some approach to a limit would be seen using PTB II 
\cite{PTB94}, which larger and more consistent for annotation than PTB I.  
As shown in Figure~\ref{fig.rule.growth.compare}, however, 
the rule number growth continues
unabated even after more than 1 million part-of-speech tokens have 
been processed.

\section{Rule Growth and Partial Bracketting}
\label{sec.partiality}

Why should the set of rules continue to grow in this way? Putting 
aside the possibility that natural languages do not have 
finite rule sets, we can think of two possible answers.  First, it 
may be that the full ``underlying grammar'' is much larger than the 
rule set that has so far been produced, requiring a much larger 
tree-banked corpus than is now available for its extraction.  If this were 
true, then the outlook would be bleak for achieving near-complete 
grammars from treebanks, given the resource demands of producing 
such resources.  However, the radical incompleteness of grammar that this 
alternative implies seems incompatible with the promising parsing 
results that Charniak reports \cite{Charniak96}.

A second answer is suggested by the presence in the extracted grammar
of rules such as (\ref{eqn.partial-rule.1}).\footnote {PTB POS tags
are used here, i.e.  DT for determiner, CC for coordinating
conjunction (e.g `and'), NN for noun} This rule is suspicious from a
linguistic point of view, and we would expect that the text from which
it has been extracted should more properly have been analysed using
rules (\ref{eqn.partial-rule.2},\ref{eqn.partial-rule.3}),
i.e. as a coordination of two simpler NPs.
\begin{equation}\it
NP\ \rightarrow\ DT\ NN\ CC\ DT\ NN
\label{eqn.partial-rule.1}
\end{equation}
\begin{equation}\it
NP\ \rightarrow\ NP\ CC\ NP
\label{eqn.partial-rule.2}
\end{equation}
\begin{equation}\it
NP\ \rightarrow\ DT\ NN
\label{eqn.partial-rule.3}
\end{equation}
\noindent Our suspicion is that this example reflects a 
widespread phenomenon of {\it partial bracketting\/} within the 
PTB.  Such partial bracketting will arise during the hand-parsing of texts, 
with (human) parsers adding brackets where they are confident that 
some string forms a given constituent, but leaving out many brackets 
where they are less confident of the constituent structure of the text.  
This will mean that many rules extracted from the corpus will be 
`flatter' than they should be, corresponding properly to what should 
be the result of using several grammar rules, showing only the top 
node and leaf nodes of some unspecified tree structure (where the 
`leaf nodes' here are category symbols, which may be nonterminal).  
For the example above, a tree structure that should properly have been
given as (\ref{trees-i}), has instead received only the partial 
analysis (\ref{trees-ii}), from the flatter 
`partial-structure' rule (\ref{eqn.partial-rule.1}).
\begin{equation}
\mbox{\hspace*{5mm} i. \hspace*{-10mm}
\Tree [.NP [.NP DT NN ] CC [.NP DT NN ] ] }
\label{trees-i}
\end{equation}
\begin{equation}
\mbox{ii. \hspace*{-10mm} \Tree [.NP DT NN CC DT NN ] }
\label{trees-ii}
\label{trees}
\end{equation}

\section{Grammar Compaction}
\label{sec.compaction}

The idea of partiality of structure in treebanks and their grammars
suggests a route by which treebank grammars may be reduced in size,
or {\it compacted\/} as we shall call it, by the elimination of
partial-structure rules.  A rule that may be eliminable as a
partial-structure rule is one that can be `parsed' (in the familiar
sense of context-free parsing) using {\it other\/} rules of the
grammar.  For example, the rule (\ref{eqn.partial-rule.1}) can be
parsed using the rules
(\ref{eqn.partial-rule.2},\ref{eqn.partial-rule.3}), as the structure 
(\ref{trees-i}) demonstrates.  Note that, although a partial-structure
rule should be parsable using other rules, it does not follow that
every rule which is so parsable is a partial-structure rule that
should be eliminated.  There may be linguistically correct rules
which can be parsed.  This is a topic to which we
will return at the end of the paper (Sec.\ \ref{sec.lvalid}).  For
most of what follows, however, we take the simpler path of assuming
that the parsability of a rule is not only necessary, but also
sufficient, for its elimination.

Rules which can be parsed using other rules in the grammar are {\it 
redundant\/} in the sense that eliminating such a rule will {\em 
never} have the effect of making a sentence unparsable that could 
previously be parsed.\footnote{Thus, wherever a sentence has a 
parse $P$ that employs the parsable rule $R$, it also has a 
further parse that is just like $P$ except that any use of $R$ is 
replaced by a more complex substructure, i.e. a parse of $R$.} 

The algorithm we use for compacting a grammar is straightforward.  A 
loop is followed whereby each rule $R$ in the grammar is addressed in 
turn.  If $R$ can be parsed using other rules (which 
have not already been eliminated) then $R$ is deleted (and the grammar 
{\it without\/} $R$ is used for parsing further rules).  Otherwise $R$ 
is kept in the grammar. The rules that remain when all rules have 
been checked constitute the compacted grammar. 

An interesting question is whether the result of compaction is 
independent of the order in which the rules are addressed.  In 
general, this is not the case, as is shown by the following rules, of 
which (\ref{eq.rule.1}) and (\ref{eq.rule.2}) can each be used to 
parse the other, so that whichever is addressed first will be 
eliminated, whilst the other will remain.
\begin{equation}\it
B\ \rightarrow\ C
\end{equation}
\begin{equation}\it
C\ \rightarrow\ B
\end{equation}
\begin{equation}\it
A\ \rightarrow\ B\ B
\label{eq.rule.1}
\end{equation}
\begin{equation}\it
A\ \rightarrow\ C\ C
\label{eq.rule.2}
\end{equation}
\noindent Order-independence can be shown to hold for 
grammars that contain no {\it unary\/} or {\it epsilon\/} (`empty') 
rules, i.e.  rules whose righthand sides have one or zero elements.  
The grammar 
that we have extracted from PTB II, and 
which is used in the compaction experiments reported in the next 
section, is one that excludes such rules.
Unary and sister rules were collapsed with
the sister nodes, e.g. the structure {\tt (S (NP -NULL-) (VP VB (NP (QP
...))) .)} will produce the following rules: {\tt S -> VP .},
{\tt VP -> VB QP} and {\tt QP -> ...}.\footnote{See \cite{Gaizauskas95} for
  discussion.}
For further discussion, and for the proof of the order independence 
see \cite{Krotov98memo}.


\section{Experiments}
\label{sec.experiments}

We conducted a number of compaction experiments: 
\footnote{For these experiments, we used two parsers: Stolcke's
BOOGIE \cite{Stolcke95:earley-cl} and Sekine's Apple
Pie Parser \cite{Sekine95}.}  first, the complete grammar was parsed 
as described
in Section~\ref{sec.compaction}.  Results exceeded our expectations:
the set of 17,529 rules reduced to only 1,667 rules, a better than
90\% reduction.

\begin{figure}
\epsfxsize=3in
\epsfysize=1.5in
\epsfbox{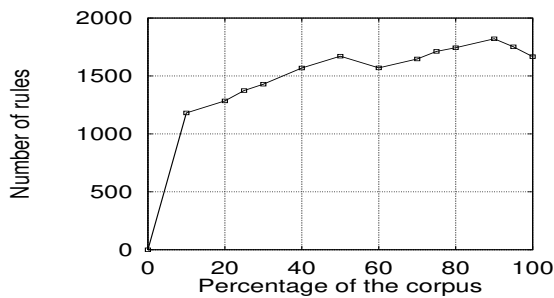}
\caption{Compacted Grammar Size}
\label{fig.comp.gram}
\end{figure}

To investigate in more detail how the compacted grammar grows, we
conducted a third experiment involving a {\it staged\/} compaction of
the grammar. Firstly, the corpus was split into 10\% chunks (by number
of files) and the rule sets extracted from each. 
The staged compaction proceeded as follows: the rule set of
the first 10\% was compacted, and then the rules for the next 10\%
added and the resulting set again compacted, and then the rules for
the next 10\% added, and so on.  Results of this experiment
are shown in Figure~\ref{fig.comp.gram}.


\noindent
At 50\% of the corpus processed the compacted grammar size actually 
exceeds the 
level it reaches at 100\%, and then the overall grammar size starts to 
go down as well 
as up.  This reflects the fact that new rules are either redundant, 
or make ``old'' rules redundant, so that the compacted grammar size 
seems to 
approach a limit.  



\section{Retaining Linguistically Valid Rules}
\label{sec.lvalid}

Even though parsable rules are redundant in the sense that has been
defined above, it does not follow that they should always be removed.
In particular, there are times where the flatter structure allowed by
some rule may be more linguistically correct, rather than simply a
case of partial bracketting.  Consider, for example, the (linguistically
plausible) rules
(\ref{eqn.red-rule.1},\ref{eqn.red-rule.2},\ref{eqn.red-rule.3}).
Rules (\ref{eqn.red-rule.2}) and (\ref{eqn.red-rule.3}) can be used to
parse (\ref{eqn.red-rule.1}), but the latter should not be eliminated, as
there are cases where the flatter structure it allows is more
linguistically correct.
\begin{equation}\it
VP\ \rightarrow\ VB\ NP\ PP
\label{eqn.red-rule.1}
\end{equation}
\begin{equation}\it
VP\ \rightarrow\ VB\ NP
\label{eqn.red-rule.2}
\end{equation}
\begin{equation}\it
NP\ \rightarrow\ NP\ PP
\label{eqn.red-rule.3}
\end{equation}
\begin{equation}
\mbox{\hspace*{5mm} i. \hspace*{-10mm}
\Tree [.VP VB [.NP NP PP ] ] }
\mbox{ii. \hspace*{-10mm} \Tree [.VP VB NP PP ] }
\label{trees-ling}
\end{equation}

We believe that a solution to this problem can be found by exploiting
the data provided by the corpus.  Frequency of
occurrence data for rules which have been collected from the
corpus and used to assign probabilities to rules, and hence to
the structures they allow, so as to produce a {\it
probabilistic\/} context-free grammar for the rules. Where a parsable 
rule is
correct rather than merely partially bracketted, we then expect 
this fact to be
reflected in rule and parse probabilities (reflecting the occurrence
data of the corpus), which can be used to decide when a rule that {\em
may} be eliminated {\em should} be eliminated.  In particular, a rule
should be eliminated only when the more complex structure allowed by
other rules is more probable than the simpler structure that the rule
itself allows. 

We developed a linguistic compaction algorithm employing the ideas
just described.  However, we cannot present it here due to the space 
limitations.
The preliminary results of our experiments are presented in
Table~\ref{tab:ling}.  Simple thresholding (removing rules that only 
occur once) was also to achieve 
the maximum compaction ratio.  For labelled as well as unlabelled 
evaluation of the
resulting parse trees we used the {\tt evalb} software by Satoshi 
Sekine.  See \cite{Krotov98memo} for the complete presentation of our
methodology and results.


\begin{table*}[htbp]
  \begin{center}
    \leavevmode
    \begin{tabular}{|p{1.5in}|l|l|l|l|l|}
    
      \hline 
      & Full & 
      Simply thresholded &  Fully compacted &
      \multicolumn{2}{|c|}{Linguistically compacted} \\

      & & & & Grammar 1 & Grammar 2 \\

      \hline

  \multicolumn{6}{|c|}{Labelled evaluation} \\

      \hline 
  Recall    & 70.55\% & 70.78\% & 30.93\% & 71.55\% & 70.76\% \\
      \hline
  Precision & 77.89\% & 77.66\% & 19.18\% & 72.19\% & 77.21\% \\
      \hline

  \multicolumn{6}{|c|}{Unlabelled evaluation} \\

      \hline 
  Recall    & 73.49\% & 73.71\% & 43.61\% & 74.72\% & 73.67\% \\
      \hline
  Precision & 81.44\% & 80.87\% & 27.04\% & 75.39\% & 80.39\% \\
      \hline

  Grammar size & 15,421 & 7,278 & 1,122 & 4,820 & 6,417 \\
      \hline
  reduction (as \% of full) & 0\% & 53\% & 93\% & 69\% & 58\% \\ 
      \hline

    \end{tabular}
    \caption{Preliminary results of evaluating the grammar compaction method}
    \label{tab:ling}
  \end{center}
\end{table*}

As one can see, the fully compacted grammar yields poor recall and
precision figures.  This can be because collapsing of the rules often
produces too much substructure (hence lower precision figures) and 
also because many longer rules in fact encode valid linguistic information.
However, linguistic compaction combined with simple thresholding
achieves a 58\% reduction without 
any loss in performance, and 69\% reduction even yields higher recall.



\section{Conclusions}


We see the principal results of our work to be the following:

\begin{itemize}
\item the result showing continued square-root 
growth in the rule set extracted
from the PTB II;

\item the analysis of the source of this continued growth in terms of
{\it partial bracketting\/} and the justification this provides for compaction
via rule-parsing;

\item the result that the compacted rule set {\it does\/} approach a 
limit
at some point during staged rule extraction and compaction, after a
sufficient amount of input has been processed;

\item that, though the fully compacted grammar produces lower 
parsing performance than the extracted grammar, 
a 58\% reduction (without loss) can still be achieved by using linguistic
  compaction, and 69\% reduction yields a gain in recall, but a loss in
  precision. 

\end{itemize}

\noindent
The latter result in particular provides further support for
the possible future utility of the compaction algorithm.  Our method is similar
to that used by Shirai \cite{Shirai95}, but the principal
differences are as follows.  First, their algorithm does not employ
full context-free parsing in determining the redundancy of rules,
considering instead only direct composition of the rules (so that
only parses of depth 2 are addressed).  We proved that the
result of compaction is independent of the order in which the rules in
the grammar are parsed in those cases involving 'mutual parsability'
(discussed in Section~\ref{sec.compaction}), but Shirai's algorithm will
eliminate both rules so that coverage is lost.  Secondly, it is
not clear that compaction will work in the same way for English as it
did for Japanese.








\bibliographystyle{acl}

\begin{small}
\bibliography{bib/new}
\end{small}


\end{document}